\documentclass[10pt,twocolumn,letterpaper]{article}

\usepackage[pagenumbers]{cvpr}              
\usepackage{booktabs}
\usepackage{multirow}
\usepackage{multicol}
\definecolor{cvprblue}{rgb}{0.21,0.49,0.74}
\usepackage[pagebackref,breaklinks,colorlinks,allcolors=cvprblue]{hyperref}

\def\paperID{****} 
\def\confName{CVPR}
\def\confYear{2026}

\title{PixelRush: Ultra-Fast, Training-Free High-Resolution Image Generation via One-step Diffusion}

\definecolor{mydarkblue}{rgb}{0,0.08,1}
\definecolor{mydarkgreen}{rgb}{0.02,0.6,0.02}
\definecolor{myred}{rgb}{1.0,0.0,0.0}
\definecolor{myred2}{rgb}{0.7,0.1,0.1}
\definecolor{mydarkblue2}{rgb}{0.05,0.1,0.7}
\definecolor{mypurple}{rgb}{111,0,255}
\definecolor{mypurple2}{rgb}{111,0,111}

\usepackage[normalem]{ulem}

\author{Hong-Phuc Lai, Phong Nguyen, Anh Tran\\
Qualcomm AI Research\footnotemark[1]\\
{\tt\small \{phuclai, phongnh, anhtra\}@qti.qualcomm.com}
}

\begin{document}

\twocolumn[{
\maketitle
\begin{center}
    \captionsetup{type=figure}
    \includegraphics[width=.99\textwidth]{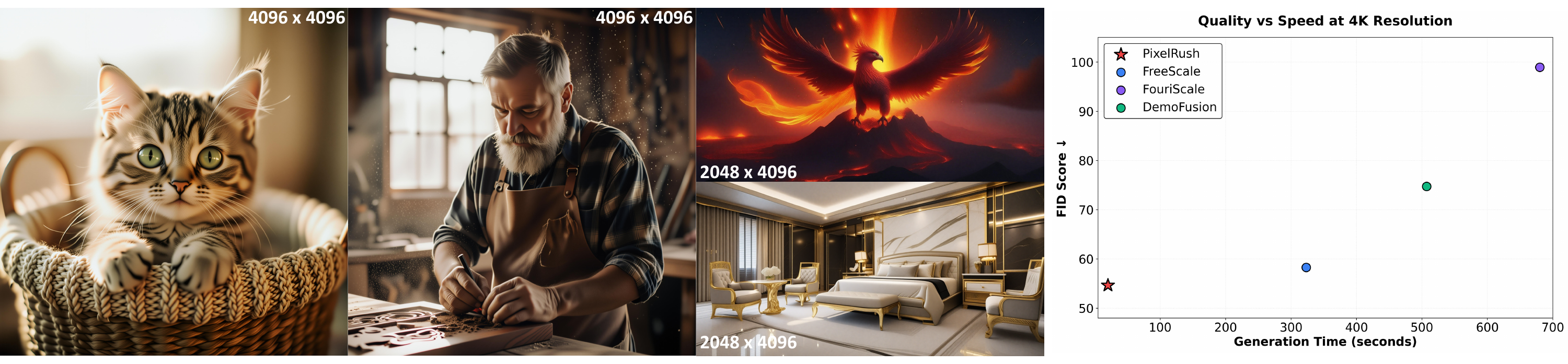}
    \captionof{figure}{\textbf{Unlocking High Resolution with PixelRush.} Our tuning-free method leverage the pretrained text-to-image models to generate high-fidelity, high-resolution images. In these examples, we extend the base model SDXL~\cite{podell2023sdxl} to generate $4K$ resolution images. Each image was produced on a single A100-40GB GPU in under 20 seconds, demonstrating state-of-the-art quality and efficiency. Best viewed \textbf{ZOOMED-IN}.}
\end{center}
}]
\renewcommand{\thefootnote}{\fnsymbol{footnote}}
\footnotetext[1]{Qualcomm Vietnam Company Limited.  Qualcomm AI Research is an initiative of Qualcomm Technologies, Inc.}
\renewcommand{\thefootnote}{\arabic{footnote}}

\begin{abstract}
Pre-trained diffusion models excel at generating high-quality images but remain inherently limited by their native training resolution. Recent training-free approaches have attempted to overcome this constraint by introducing interventions during the denoising process; however, these methods incur substantial computational overhead, often requiring more than five minutes to produce a single 4K image.
In this paper, we present PixelRush, the first tuning-free framework for practical high-resolution text-to-image generation. Our method builds upon the established patch-based inference paradigm but eliminates the need for multiple inversion and regeneration cycles. Instead, PixelRush enables efficient patch-based denoising within a low-step regime. To address artifacts introduced by patch blending in few-step generation, we propose a seamless blending strategy. Furthermore, we mitigate over-smoothing effects through a noise injection mechanism. PixelRush delivers exceptional efficiency, generating 4K images in approximately 20 seconds representing a 10$\times$ to 35$\times$ speedup over state-of-the-art methods while maintaining superior visual fidelity. Extensive experiments validate both the performance gains and the quality of outputs achieved by our approach.
\end{abstract}

\section{Introduction}
Diffusion models have significantly advanced the field of visual synthesis~\cite{ho2020denoising, song2020score,dhariwal2021diffusion}, generating images with unprecedented photorealism and diversity. However, these models are typically trained on datasets with fixed, constrained resolutions (e.g., $1024\times1024$ for SDXL~\cite{podell2023sdxl}, $512\times512$ for SD2.1~\cite{rombach2022high}). Consequently, attempting to generate images that exceed these native resolutions at inference time often leads to significant quality degradation and structural artifacts~\cite{qiu2025freescale, du2024demofusion, vontobel2025hiwave}. A straightforward solution is to fine-tune the model on the target high-resolution data. This approach, while straightforward, is often impractical due to three major obstacles: (i) the scarcity of large-scale, high-resolution training datasets, (ii) the prohibitive computational resources required for training on such data, and (iii) the inflexibility of the resulting models, which are constrained to their specific fine-tuning resolution~\cite{xie2025sana, zhang2025diffusion}. This bottleneck has created a critical need for methods that can unlock high-resolution synthesis from pre-trained models without costly re-training~\cite{du2024demofusion}.

Training-free high-resolution image generation comprises two main approaches: direct inference~\cite{qiu2025freescale,huang2024fouriscale,lin2024accdiffusion} and patch-based methods~\cite{vontobel2025hiwave, du2024demofusion}. Direct inference methods operate on the entire high-resolution latent. ScaleCrafter~\cite{he2023scalecrafter} modifies the dilation rate of convolutional layers to enlarge the receptive field, which mitigates object repetition artifacts. FreeScale~\cite{qiu2025freescale} builds on this by integrating local information to enhance high-frequency components. However, these methods face two significant limitations: (i) their manipulation of the frequency domain can introduce unnatural textures~\cite{huang2024fouriscale,lin2024accdiffusion} and (ii) their high memory footprint, which scales with the latent size, typically limits them to a maximum of $8K$ resolution~\cite{qiu2025freescale}. The patch-based paradigm was proposed to address this peak memory constraint, enabling the synthesis of images at resolutions exceeding $8K$~\cite{tragakis2024one}. Methods like DemoFusion~\cite{du2024demofusion} partition the large latent into smaller, overlapping patches, processing each at a size compatible with the pre-trained model's native resolution. While this effectively removes the memory bottleneck, it shares a critical drawback with direct inference methods: prohibitively slow inference speeds. The standard procedure for both involves taking a coarse, upscaled image, perturbing it to full Gaussian noise (at $t=T$), and then executing a long, multi-step reverse diffusion process (e.g., 50 steps) to synthesize high-frequency details. This full denoising trajectory results in impractical generation times, requiring several minutes for a $4K$ image and over an hour for an $8K$ image~\cite{qiu2025freescale, vontobel2025hiwave}.

Attempts to address this inference bottleneck have been limited and met with marginal success. CutDiffusion~\cite{lin2024cutdiffusion}, proposes a two-stage process that reduces the number of patches in the first stage to cut computation. However, this yields only minor speed-ups and often comes at the cost of quality. Another approach, LSNR~\cite{jeong2025latent}, requires training a plug-in latent super-resolution module. While it can reduce the required denoising steps (e.g., from $50$ to $30$ when combined with DemoFusion~\cite{du2024demofusion}), this gain is modest and still operates firmly within a multi-step regime. This incompatibility with fast, few-step sampling remains the primary barrier to practical applications.

To finally make high-resolution synthesis practical, we propose PixelRush, a tuning-free, patch-based strategy that is the first to be successfully designed for the few-step sampling regime. Specifically, our analysis of current multi-step pipelines reveals that they follow a hierarchical generation process in the frequency domain: low-frequency global structures are formed early in the reverse process, while high-frequency details are synthesized in later stages. This leads to our key insight: for a refinement task, executing the full reverse process is computationally redundant. We are the first to propose a partial inversion technique that  focusing computation exclusively on the synthesis of fine-grained details. This approach is synergistic with few-step diffusion models, as their update steps can generate all necessary details within short, truncated trajectory. The result is a pipeline that is orders of magnitude faster than the current state-of-the-art. 

The proposed paradigm shift, however, introduces a critical and unique challenge: standard patch blending methods, which work in multi-step processes, fail in a low-step context and produce severe boundary artifacts. To solve this, we first propose a novel patch blending algorithm, inspired by image feathering, that seamlessly stitches patches even when using a minimal number of steps. Second, to counteract the oversmoothing typically observed in few-step generation, we introduce a noise injection technique that effectively preserves high-frequency details. 

The impact of our approach is dramatic. PixelRush achieves an unprecedented 10$\times$ to 35$\times$ acceleration over current multi-step SOTA methods~\cite{du2024demofusion, qiu2025freescale}, while delivering superior quality, for example, improve the SOTA FID score from 52.87 to 50.13. This efficiency makes our method the first to generate $8K$ images in under 100 seconds on a single A100-40GB GPU. Our contributions are summarized as follows:
\begin{itemize}
    \item     We propose PixelRush, a novel and efficient training-free pipeline for high-resolution image generation that, for the first time, makes few-step sampling viable for this task. 

   \item  We introduce a partial inversion strategy, a smooth blending algorithm, and a noise injection technique as key components to achieve both speed and quality, resolving the core bottlenecks of prior art.

    \item We conduct extensive experiments that demonstrate that PixelRush achieves state-of-the-art quantitative and qualitative results, with an unprecedented acceleration $10\times$ to $35\times$ over existing methods.
\end{itemize}

\section{Related work}
\paragraph{Advances in Diffusion-Based Visual Synthesis.} Diffusion models have advanced the field of image generation by producing images of remarkable quality~\cite{ho2020denoising, song2020score, dhariwal2021diffusion}. Latent diffusion models~\cite{rombach2022high} built on a U‑Net backbone improve efficiency by operating in a compressed latent space, enabling complex tasks such as text‑to‑image synthesis (e.g., Stable Diffusion, SDXL)~\cite{podell2023sdxl}. While U‑Net architectures dominated early diffusion work, recent studies show that diffusion transformers scale more effectively~\cite{peebles2023scalable, ma2024sit}, and current state‑of‑the‑art diffusion systems increasingly adopt transformer architectures~\cite{chen2023pixart, xie2025sana, esser2024scaling}. Building on these foundation models, our work focuses on intervening in the inference process of pretrained models such as SDXL~\cite{podell2023sdxl} to enhance their ability to generate high‑resolution images.\par
A key drawback of diffusion is its reliance on slow, multi‑step inference. This cost can be mitigated by applying distillation methods~\cite{wang2023prolificdreamer, nguyen2024swiftbrush, song2023consistency,yin2024one} to distill a multi‑step diffusion teacher into a one‑step or few‑step student without degrading sample quality. We leverage the efficiency of few‑step models to accelerate our refinement pipeline while preserving high‑fidelity generation.\par 
\paragraph{Tuning-free High-resolution Synthesis.} MultiDiffusion~\cite{bar2023multidiffusion} and DemoFusion~\cite{du2024demofusion} partition high‑resolution latents into patches sized to the pretrained model’s native resolution, process patches independently, and stitch them using average blending technique. While reducing the memory footprint, these pipelines frequently exhibit object repetition. FouriScale~\cite{huang2024fouriscale} attributes this failure to frequency‑domain misalignment across resolutions, and both FouriScale~\cite{huang2024fouriscale} and FreeScale~\cite{qiu2025freescale} introduce frequency interventions to counteract it. Although repetition is reduced, direct manipulation in the frequency domain often yields unnatural textures and additional high‑frequency artifacts. All these methods typically initiate the upsampling stage by perturbing the coarse, target‑resolution latent with Gaussian random noise. In contrast, HiWave~\cite{vontobel2025hiwave} injects DDIM‑inverted~\cite{song2020denoising} noise, which preserves structural information and also helps mitigate object repetition.\par
Another major limitation of these methods is their slow inference: all rely on the multi‑step reverse process~\cite{du2024demofusion,bar2023multidiffusion,huang2024fouriscale,qiu2025freescale, he2023scalecrafter,vontobel2025hiwave} of pretrained diffusion models. CutDiffusion~\cite{lin2024cutdiffusion} attempts to reduce runtime by processing fewer patches, but the improvement is only incremental. 
In contrast, our method addresses both limitations with a novel partial inversion pipeline that leverages a few‑step diffusion model. By initiating the upsampling stage from a structure‑preserving latent, it inherently mitigates small object repetition. This design also enables generation at resolutions of $8K$ and beyond on a single GPU, while dramatically shortening inference time compared to prior patch‑based methods.

\begin{figure}[t]
\centering
\includegraphics[width = 0.49\textwidth]{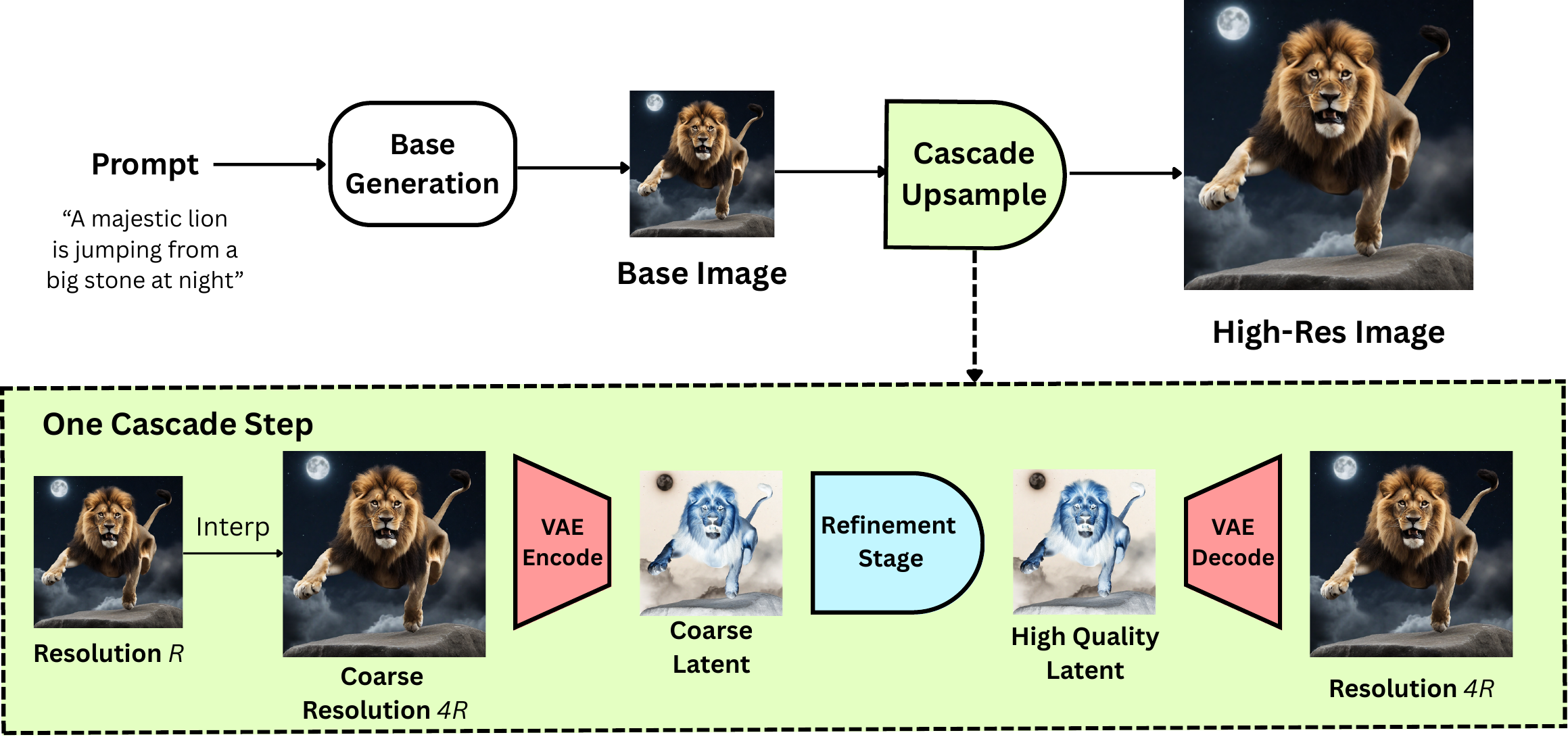}
\caption{\textbf{An overview of two-stage system with for high-resolution generation with Cascade Upsampling.} (a) \textit{Two-Stage System.} A base diffusion model generates a low-resolution, base image. This image then goes into a cascade upsampling process to progressively upscale to target resolution. (b) \textit{The Cascade Step.} Each cascade step doubles the height and width of an input image. First, the initial image at resolution $R$ is upscaled to $4R$ via interpolation in pixel space, creating a coarse image. This coarse image is then encoded by VAE encoder to obtain a coarse latent. This coarse latent is enhanced with high-frequency details synthesized in our Refinement Stage, yield a high-quality latent. Finally, this refined latent is decoded back to the pixel space, producing a sharp, high-fidelity image at resolution $4R$.}

\label{fig:pipeline}
\vspace{-3mm}
\end{figure}

\paragraph{Super-Resolution.} Image Super-Resolution (SR) is a classical computer vision task focused on reconstructing a high-resolution (HR) image from a given, authentic low-resolution (LR) input. Studies on image super-resolution have a long history, from classical interpolation~\cite{yue2016image, van2006image} to modern deep learning approaches using convolutional networks, generative adversarial networks~\cite{chen2024ntire, dong2014learning, wang2020deep}, and more recently, diffusion models~\cite{li2022srdiff, wu2024one} trained specifically for the SR task. It is important to distinguish our high‑resolution generation task from classical super‑resolution. In super‑resolution, the low‑resolution image is a fixed, real‑world input to be faithfully restored. In our pipeline, the low‑resolution image is an intermediate product that provides a structural foundation. Consequently, our aim is not pixel‑perfect reconstruction but coherent novel synthesis: produce a high‑resolution image that preserves the intermediate result’s global layout and semantics while introducing plausible high‑frequency details consistent with the original text prompt.
\begin{figure*}[t]
\centering
\includegraphics[width = 0.98\textwidth]{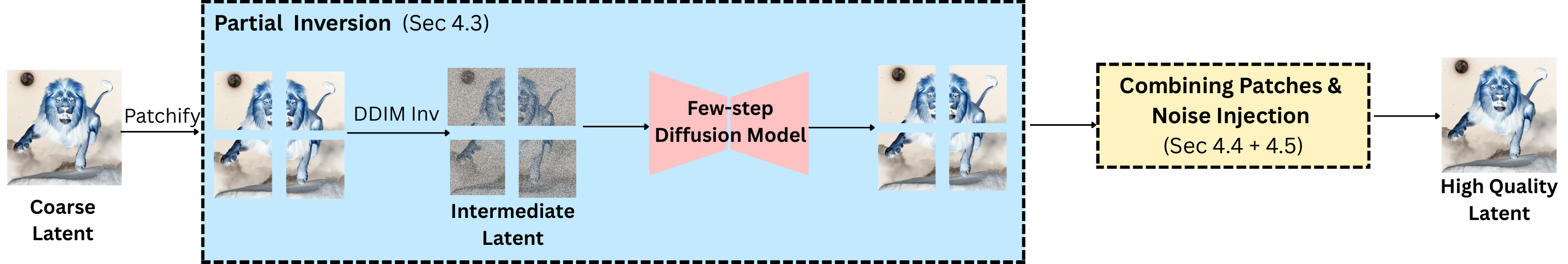}
\caption{\textbf{The PixelRush Refinement Stage.} Our refinement stage takes a coarse latent as input and first divides it into overlapping patches. These patches pass through proposed partial inversion few-step pipeline (Sec.~\ref{subsec:speed-up}), where DDIM inversion maps each patch to an intermediate noisy latent. A few-step diffusion model then refines these latents, synthesizing high-frequency details. Finally, the refined patches are processed by our Gaussian Filter Patches Blending \& Noise Injection module (Sec.~\ref{subsec:feathering} + Sec.~\ref{subsec:noise_inject}) to produce a seamless, high-quality latent.}

\label{fig:pixelrush}
\vspace{-5mm}
\end{figure*}
\section{Background}
\subsection{Latent Diffusion Models}
\label{subsec:diffusion}
Latent Diffusion Models~\cite{rombach2022high} reduce computation by operating in a compressed latent space. Given an image $x$, an encoder $\mathcal{E}$ produces the latent $\mathbf{z}_{0}=\mathcal{E}(x)$. The latent diffusion models first gradually perturb the clean latent $\mathbf{z}_{0}$ with gaussian noise (forward process) and learn a network to invert that corruption (reverse process). The forward process adds noise to $\mathbf{z}_{0}$ over $T$ steps to produce an approximately gaussian noise $\mathbf{z}_{T}$. By the Markov property, the probability distribution of an intermediate latent $\mathbf{z}_{t}$ given $\mathbf{z}_{0}$ is
\begin{equation}
    q(\mathbf{z}_t|\mathbf{z}_0) = \mathcal{N}(\mathbf{z}_t;\sqrt{\bar{\alpha}_t}\mathbf{z}_0, (1-\bar{\alpha}_t)\mathbb{I})
\end{equation}
where $\bar{\alpha}_{t}$ is a schedule‑dependent hyperparameter. The reverse process learns a denoising network $\epsilon_{\theta}(\mathbf{z}_{t},t)$ and parameterizes the conditional distribution 
\begin{equation}
    p_\theta(\mathbf{z}_{t-1}|z_t) = \mathcal{N}(\mathbf{z}_{t-1};\mu_\theta(z_t, t), \Sigma_{\theta}(\mathbf{z}_t, t))
\end{equation}
with $\mu_\theta(\mathbf{z}_t, t)$ and $ \Sigma_{\theta}(\mathbf{z}_t, t)$ computed from $\epsilon_{\theta}(\mathbf{z}_{t},t)$.
\subsection{DDIM Inversion}
 Denoising Diffusion Implicit Model (DDIM~\cite{song2020denoising}) provides a deterministic sampling path that allows for much faster sampling with fewer steps and enables a consistent mapping between the Gaussian noise space and the clean latent space. The goal of DDIM inversion is to find a latent noise vector $\mathbf{z}_T$ that, when used as the starting point for the standard DDIM sampling process, reconstructs a given source image $\mathbf{x}_0$. The deterministic DDIM inversion mapping from $\mathbf{z}_{t-1}$ to $\mathbf{z}_{t}$ is defined as:
\begin{equation}
    \label{eq:ddim_step}
    \begin{split}
        \mathbf{z}_{t} = \sqrt{\bar{\alpha}_{t}} \left( \frac{\mathbf{z}_{t-1} - \sqrt{1-\bar{\alpha}_{t-1}}\epsilon_{\theta}(\mathbf{z}_{t-1}, t)}{\sqrt{\bar{\alpha}_{t-1}}} \right) \\
                         \quad + \sqrt{1-\bar{\alpha}_{t}} \epsilon_{\theta}(\mathbf{z}_{t-1}, t),
    \end{split}
\end{equation}
where $\bar{\alpha}_t$ is the hyperparameter of the noise scheduler and $\epsilon_{\theta}$ is the denoiser network. This process can be stopped at any intermediate timestep $t$ to find a noisy latent $\mathbf{z}_{t}$ that lies on the trajectory leading back to the original latent $\mathbf{x}_0$. 
    
\section{Method}
We present the PixelRush pipeline for high-resolution image generation using pretrained diffusion models. We first provide an overview of the system architecture (Sec.~\ref{subsec:overall}), then describe our motivation and design choices to enable patch-based inference in the few-step diffusion regime (Sec.~\ref{subsec:partial_gaussian} and \ref{subsec:speed-up}), introduce our patch blending technique inspired by image feathering that eliminates artifacts (Sec.~\ref{subsec:feathering}) and finally introduce our noise injection technique to prevent oversmoothing in few-step setup (Sec.~\ref{subsec:noise_inject}).

\subsection{Overall}\label{subsec:overall}
The overall architecture of training‑free patch-based high‑resolution synthesis systems can be viewed as a two‑stage pipeline (Fig.~\ref{fig:pipeline}): a base generation stage followed by a cascade upsampling stage.

Given a prompt $c$ and a target resolution $R$, a diffusion model $\epsilon^{base}_\theta$ first generates a base latent $\mathbf{z}_{0}^n$. This latent is synthesized at the model's native resolution $n =H \times W$ with $H, W$ are latent's spatial dimension. Without loss of generalization, we can assume $R = n \times 4^{s}$ with $s \in \mathbb{N}^*$. Directly upscaling from the original resolution $n$ to $R$ can introduce visual artifacts~\cite{du2024demofusion, vontobel2025hiwave}. Therefore we adopt a cascade upsampling strategy (a single cascade step that doubles latent's spatial shape or increases resolution $4\times$ is illustrated at the bottom of Fig.~\ref{fig:pipeline}) from prior works~\cite{vontobel2025hiwave, qiu2025freescale, du2024demofusion}:
\begin{equation*}
\mathbf{z}_0^{n} \rightarrow \mathbf{z}_0^{4n} \rightarrow \dots \rightarrow \mathbf{z}_0^{R},
\end{equation*}
\begin{equation*}
\bar{\mathbf{z}}_0^{4r} = \mathbb{UP}(\mathbf{z}_0^{r}),~ \mathbf{z}_0^{4r} = \phi\big(\bar{\mathbf{z}}_0^{4r}\big) \quad \forall r = n \times 4^{i-1}, i \in [1..s],
\end{equation*}
where $\mathbf{z}_t^r$ denotes the latent with resolution $r$ at timestep $t$, $\mathbb{UP}$ denotes an upsampling operation that moves the current latent to the next resolution level, $\bar{\mathbf{z}}_0^{r}$ denotes the coarse latent containing primarily low‑frequency (missing high‑frequency) content, $\mathbf{z}_0^{r}$ denotes the fine-grained detail latent and $\phi$ denotes the refinement stage. The upsampling operation $\mathbb{UP}$ can be performed in latent space $\mathbb{UP}_{\mathrm{latent}}(\mathbf{z}_0^{r})=\mathrm{Interpolate}(\mathbf{z}^{r},2)$, or in pixel space $\mathbb{UP}_{\mathrm{pixel}}(\mathbf{z}^{r})=\mathcal{E}\big(\mathrm{Interpolate}(\mathcal{D}(\mathbf{z}^{r}),2)\big)$, where $\mathrm{Interpolate}(\cdot,k)$ is a standard upsampling operator, e.g., bilinear or bicubic upsampling, with the scale factor $k$ and $\mathcal{D}$ and $\mathcal{E}$ are the decoder and encoder that map between latent and pixel spaces. Following recent work~\cite{qiu2025freescale, vontobel2025hiwave}, we perform the upsampling operation in pixel space rather than latent space. This approach has been shown to prevent the introduction of artifacts that can arise from direct manipulation of the latent representation.

In prior methods~\cite{du2024demofusion, vontobel2025hiwave, qiu2025freescale}, the refinement stage first perturbs the coarse latent $\bar {\mathbf{z}}_{0}^{r}$ to a (approximately) full Gaussian $\bar{\mathbf{z}}_{T}^{r}$ and then runs a reverse diffusion denoising process to add fine‑grained details and obtain $\mathbf{z}_{0}^{r}$. Crucially, all such refinement pipelines depend on a full multi‑step reverse diffusion which is the dominant source of their computational cost and the bottleneck we address in the next section.
\subsection{Is fully Gaussian noise needed?}\label{subsec:partial_gaussian}
\begin{figure}[t]
\centering
\includegraphics[width = 0.49\textwidth]{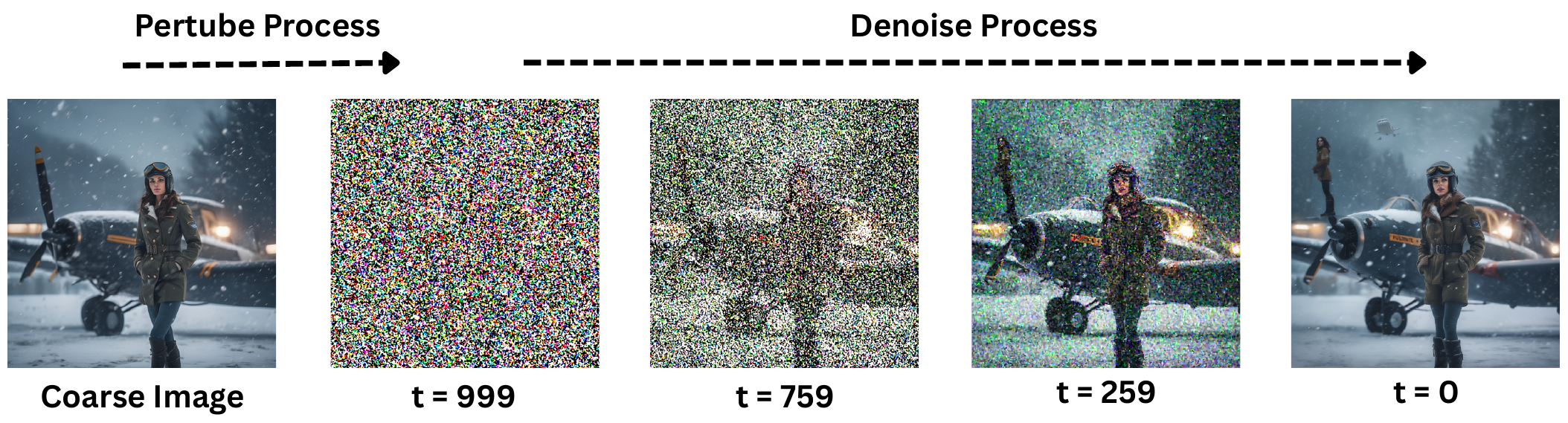}
\caption{\textbf{Training-free high-resolution pipeline synthesize images hierarchically.}} 
\vspace{-5mm}
\label{fig:sedit}
\end{figure}

It is well established that the reverse diffusion process reconstructs images hierarchically in the frequency domain, producing low‑frequency, global structure first and adding high‑frequency, fine‑grained details later~\cite{meng2021sdedit}. We find the same temporal ordering in high‑resolution generation. As illustrated in Fig~\ref{fig:sedit}, DemoFusion~\cite{du2024demofusion} follows this frequency‑wise reconstruction. This observation motivates the question: Is perturbing the latent all the way to a full Gaussian noise, i.e., $t=T$, truly necessary?

We argue that perturbing to full Gaussian noise is suboptimal because denoising steps at large timesteps are largely redundant: they mainly restore global structure, which is already present in the coarse latent ($\bar{\mathbf{z}}_0^{r}$), and they substantially increase inference time. Therefore, we propose truncating the forward perturbation and instead mapping ($\bar{\mathbf{z}}_0^{r}$) to an intermediate noise level ($\bar{\mathbf{z}}_K^{r}$) with $(K<T)$.  For example, in Fig.~\ref{fig:sedit}, we can perturb the coarse image to timestep $t = 259$ (instead of 999), and denoise it back, saving 75\% computation time. Furthermore, the quantitative justification for this design choice is presented in our component ablation study (Table~\ref{tab:ablation_components}). By replacing the full 50-step reverse process with our 15-step partial inversion scheme, we achieve a $3.7\times$ acceleration and this dramatic speedup does not come at the cost of quality. This result provides strong empirical evidence that our partial inversion strategy is both more efficient and effective for high-fidelity refinement.

This truncated strategy naturally complements few‑step diffusion models, which can make large, high‑impact updates and synthesize the required high‑fidelity details within a shortened reverse trajectory. This is the core idea behind our approach to speed up the refinement pipeline by leveraging few‑step diffusion models.


\subsection{Speed-up with few-step model}\label{subsec:speed-up}
In the refinement stage, instead of using the same multi-step diffusion model as in previous works, we can employ a few‑step sampling diffusion model $\epsilon_{\gamma}^{\mathrm{refine}}$ to further boost inference speed, as illustrated in Fig.~\ref{fig:pixelrush}.
Motivated by the observations in Sec.~\ref{subsec:partial_gaussian}, our perturbation process during refinement maps the coarse latent $\bar{\mathbf{z}}_0^{r}$ only to an intermediate noise level $\bar{\mathbf{z}}_K^{r}$ rather than to full Gaussian noise. To maximize pipeline speedup, we perform both the forward perturbation and the reverse refinement using a single step, choosing the corresponding intermediate timestep $K$ of $\epsilon_{\gamma}^{\mathrm{refine}}$. For instance, if $\epsilon_{\gamma}^{\mathrm{refine}}$ samples over 4 equally distributed time-steps, we can select $K = 249$, enabling natural one-step inversion and one-step reverse diffusion.

For the initial refinement stage, we employ deterministic DDIM inversion. While stochastic $q$-sampling is also a feasible option, we prefer this deterministic inversion to preserves the base image's structural information.
The DDIM-inverted noise retains the crucial low-frequency content from the coarse image, enabling the few-step denoiser to focus its capacity on synthesizing high-frequency details, a principle supported by prior work~\cite{vontobel2025hiwave, song2020denoising}.
\begin{figure}[t]
\centering
\includegraphics[width = 0.47\textwidth]{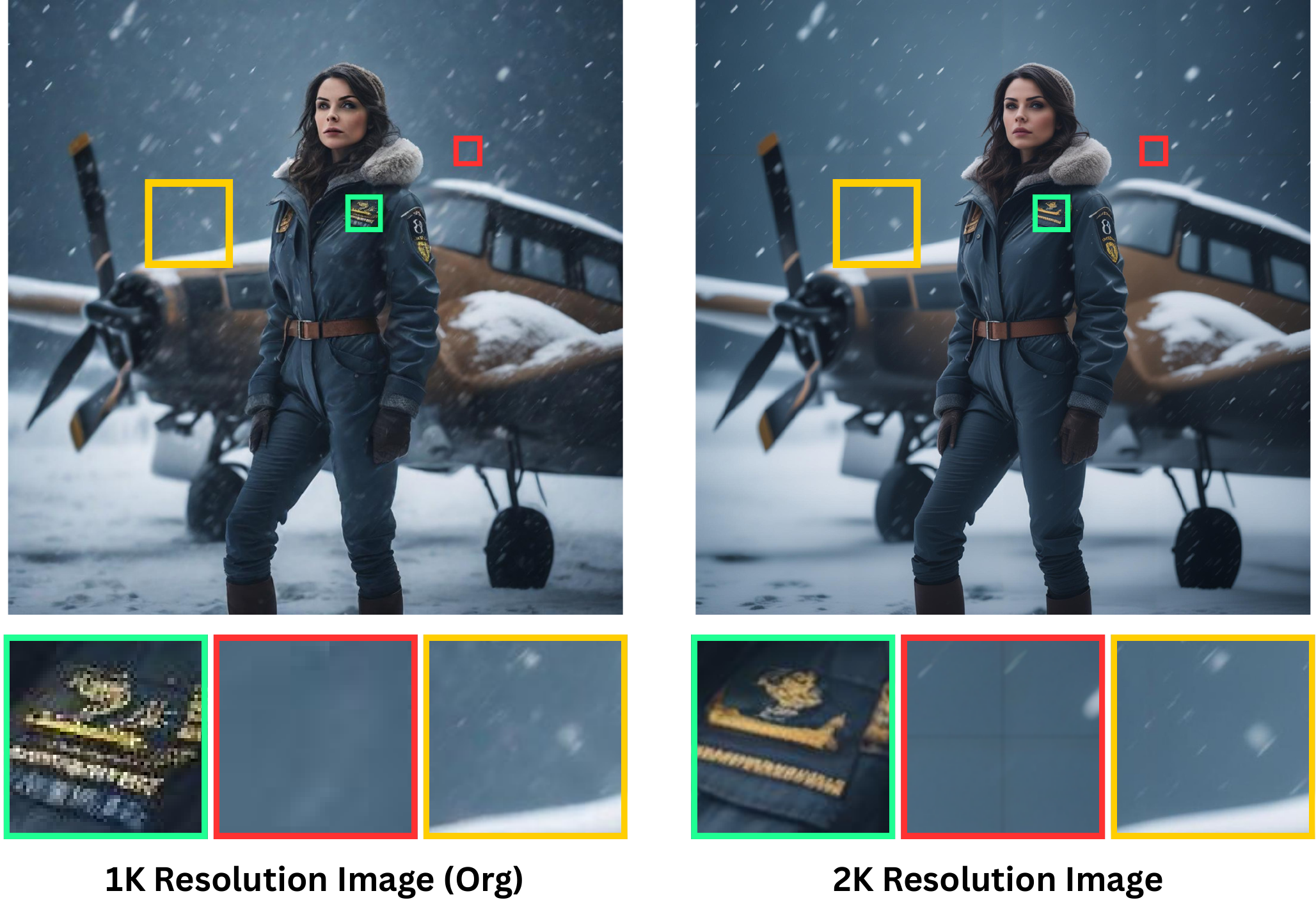}
\vspace{-2mm}
\caption{\textbf{Detail enhancement and artifact emergence.} Compared to the original 1K image, 2K image produced by our partial inversion few-step pipeline~\ref{subsec:speed-up} successfully synthesizes high-fidelity details (green box). However, this process also introduces checkerboard (red box)  and over-smoothing (yellow box) artifacts. Best viewed \textbf{ZOOMED-IN}.}
\vspace{-5mm}
\label{fig:few_step}
\end{figure}
These initial design choices yield a significant runtime speedup of approximately $10\times$ to $35\times$ and enable the synthesis of fine-grained details within the high-resolution output. However, this efficiency does not come without cost, as this baseline approach introduces noticeable checkerboard artifacts and a degree of oversmoothing in the generated images, as illustrated in Fig.~\ref{fig:few_step}. This trade-off motivates the subsequent refinements in our pipeline.

\subsection{Image feathering blending technique}\label{subsec:feathering}
Although efficient, naively equipping a patch‑based pipeline with a few‑step diffusion model produces severe artifacts at patch boundaries (see Fig.~\ref{fig:few_step}). We trace this failure to the simple averaging blending (MultiDiffusion~\cite{bar2023multidiffusion}) when applied in few‑step or one‑step regimes. Few‑step reverse processes produce large, sharp updates inside individual patches, and naively averaging overlapping regions only smooths those differences without reconciling them, yielding visible seams.

\begin{figure}[t]
\centering
\includegraphics[width = 0.49\textwidth]{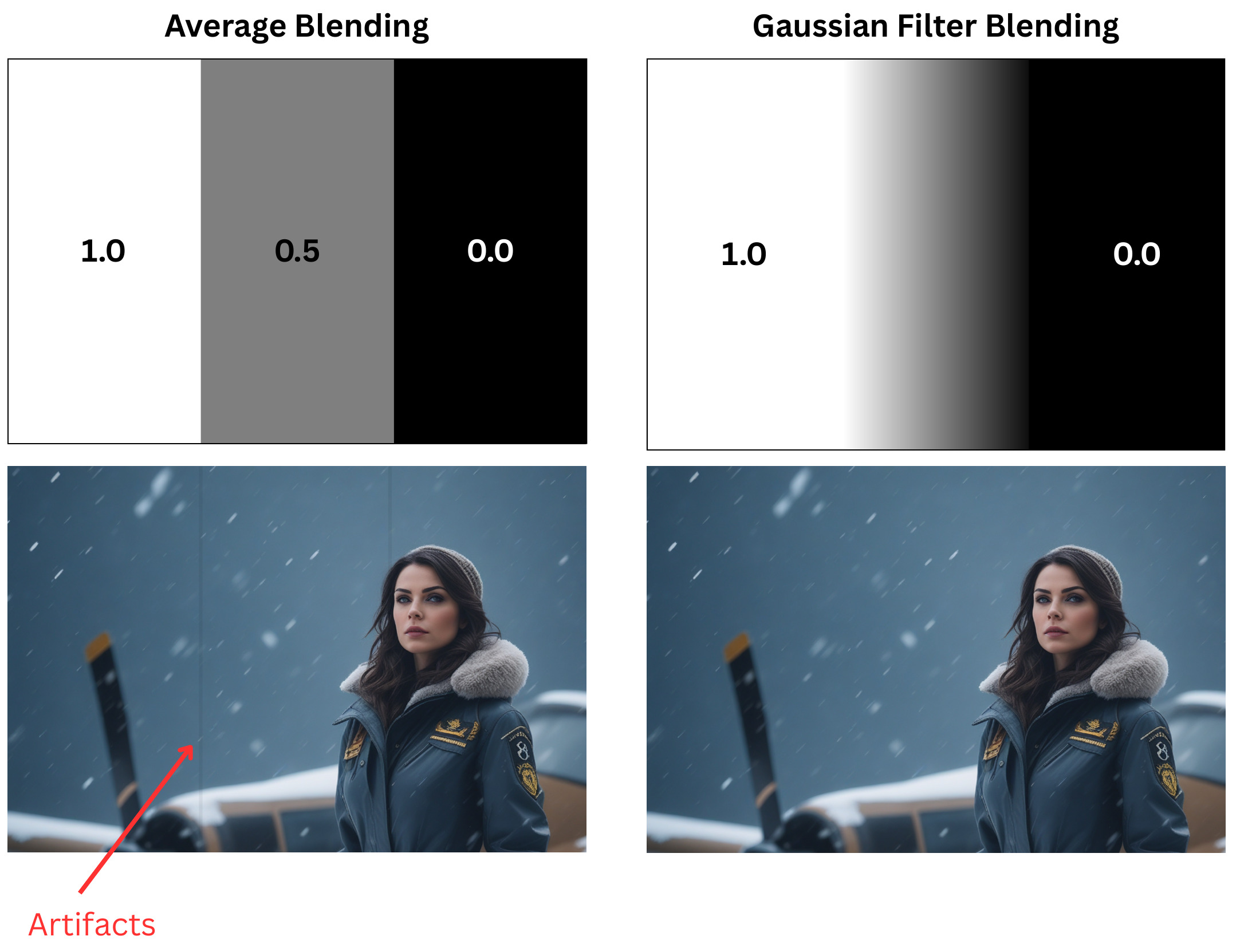}
\vspace{-5mm}
\caption{\textbf{Gaussian filter blending methods.} Our method (\textit{right}) uses a Gaussian filter to transform a discontinuous mask become smooth, continuous weighting mask. This ensures a gradual transition between patches, completely eliminating the artifact even in one-step setting. Best viewed \textbf{ZOOMED-IN}.}
\vspace{-5mm}
\label{fig:gaussian}
\end{figure}

Our idea is that when combining two overlapping patches $P_{1}$ and $P_{2}$, pixels nearer to one patch center should follow that patch more strongly, rather than applying a uniform average across the overlap. Concretely, we convolve the hard binary overlap mask with a Gaussian blur kernel to produce a smooth mask (see Fig.~\ref{fig:gaussian} for a visualization). Blending the two patches using this smooth mask, even in a single step, removes visible boundary artifacts.

\subsection{Noise injection}\label{subsec:noise_inject}
\begin{figure}[t]
\centering
\includegraphics[width = 0.47\textwidth]{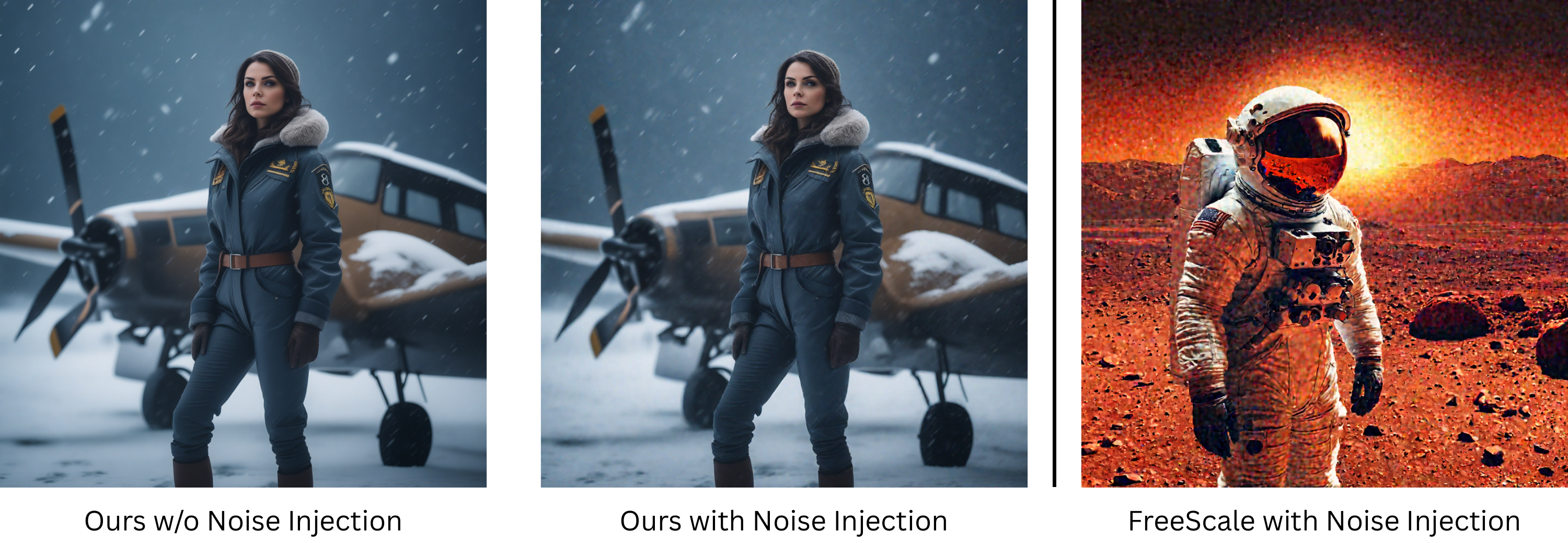}
\caption{\textbf{Our Noise Injection is a targeted fix for few-step oversmoothing.} While our method can exhibit oversmoothing (\textit{left}), the addition of proposed noise injection technique restores critical high-frequency detail (\textit{center}). Importantly, this is not an universal solution; applying it to a multi-step method like FreeScale fundamentally degrades its output and introduces noise artifacts (\textit{right}). Best viewed \textbf{ZOOMED-IN}.}
\label{fig:noise_inject}
\end{figure}

A further challenge arises when applying few-step models within our training-free, patch-based high-resolution generation framework: the resulting images often exhibit oversmoothing artifacts and a lack of fine detail, as shown in Fig.~\ref{fig:few_step}. We hypothesize that multi‑step models refine latents through many small, incremental updates that gradually synthesize high‑frequency detail. In contrast, few‑step models take larger denoising steps that may not fully recover fine‑grained details, yielding overly smooth outputs.
Prior work~\cite{xu2025temporal, song2020denoising} demonstrates that flattening the data distribution $p_\gamma(\mathbf{x})$ encourages the synthesis of higher-frequency components. Inspired by this principle, we propose a simple yet effective noise injection technique to counteract oversmoothing. We slightly augment the predicted noise $\epsilon_\gamma(\mathbf{x}, t)$ at reverse step by interpolating it with a random noise, This injects randomness which helps flatten the distribution $p_\gamma(\mathbf{x})$ and improve high-frequency details. Since the interpolation is done in latent space, as suggested by \cite{nguyen2024swiftbrush}, we employ spherical interpolation (slerp) instead of normal linear interpolation (lerp):
\begin{equation}
\epsilon'_\gamma(\mathbf{x}, t) = \operatorname{slerp}(\epsilon_\gamma(\mathbf{x}, t), \epsilon_{\mathrm{rand}}, \lambda).
\label{eq:noise_injection}
\end{equation}
where $\epsilon_{\mathrm{rand}} \sim \mathcal{N}(0, \mathbb{I})$ and $\lambda \in [0, 1)$ is an interpolation coefficient. For all our experiments, we use a fixed $\lambda = 0.95$. As demonstrated in Figure~\ref{fig:noise_inject}, this technique effectively mitigates the oversmoothing issue.

It is crucial to note that this technique is specifically tailored to counteract the oversmoothing inherent in our few-step, patch-based pipeline. When applied to standard multi-step pipelines, it can degrade performance because repeatedly steering the distribution over many steps can lead to significant error accumulation, which introduce unwanted artifacts, as shown in Figure~\ref{fig:noise_inject}. 

\section{Experiments}
\begin{figure*}[th]
\centering
\includegraphics[width = 0.99\textwidth]{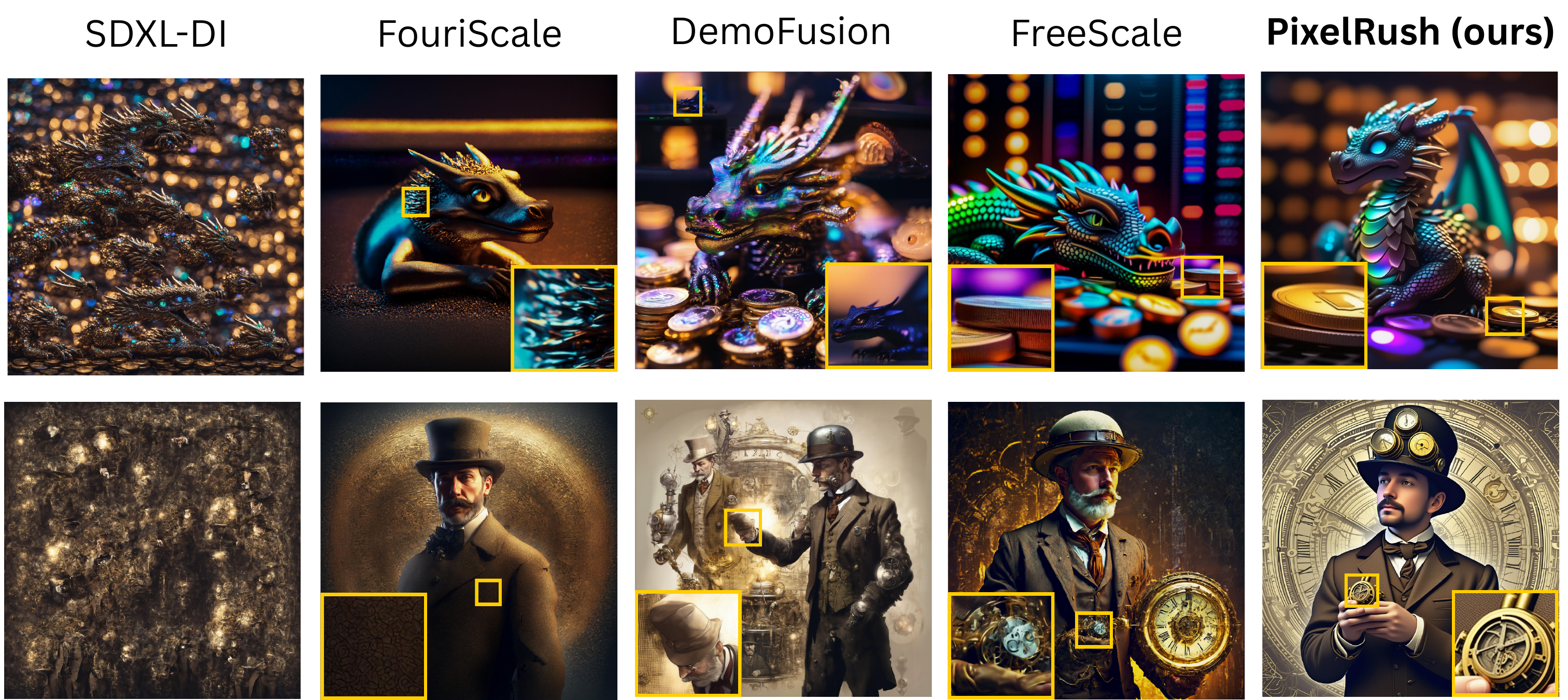}
\caption{\textbf{Qualitative Comparision.} Our PixelRush pipeline produces high-resolution images that exhibit prompt alignment and finer detail compared to existing methods. Top: $2048 \times 2048$ resolution - \textit{A tiny, palm-sized dragon with iridescent scales, curled up asleep on a pile of glowing cryptocurrency coins inside a server rack, macro photography, cinematic depth of field, bokeh lights}. Bottom: $4096\times 4096$ resolution - \textit{Victorian gentleman explorer in a tweed suit and pith helmet, discovering the glowing, mechanical core of a clockwork planet, steampunk, sense of adventure, volumetric steam}. Best viewed \textbf{ZOOMED-IN}.}
\label{fig:compare}

\end{figure*}
\begin{table*}[t]
  \centering
    \resizebox{.70\textwidth}{!}{
  \begin{tabular}{@{}l c c c c c c}
  \toprule
        
      & \multicolumn{3}{ c }{$2048\times2048$} &\multicolumn{3}{ c }{$4096\times 4096$}\\
    \cmidrule(lr){2-4}\cmidrule(lr){5-7}
& FID ($\downarrow$) & IS ($\uparrow$)  & Time (sec) & FID ($\downarrow$) & IS ($\uparrow$)  & Time (sec) \\
    \midrule
    SDXL-DI~\cite{podell2023sdxl}&73.34 &10.93 &28 &153.53 &7.32 &247\\
    FouriScale~\cite{huang2024fouriscale} &72.65 &12.31 &87 &98.97 &8.54 &680 \\ 
    DemoFusion~\cite{du2024demofusion} &68.46 &13.15 &75 &74.75 &12.57 &507\\ 
    FreeScale~\cite{qiu2025freescale} &52.87 &13.56 &53 &58.28 &13.35 &323\\ 
    PixelRush\textsuperscript{*} & \textbf{50.13} &$\textbf{14.32}$ &$\textbf{4}$ & \textbf{54.67} &$\textbf{13.75}$ &$\textbf{20}$\\
    \bottomrule
  \end{tabular}
  }
  \caption{\textbf{Quantitative Comparision.} We compare our method against state-of-the-art training-free methods for high-resolution image generation. An asterisk (*) indicates that the upsampling stage uses the SDXL-Turbo model with a single inference step. All other methods use the standard SDXL with 50 steps for this stage.}
  \vspace{-1mm}
  \label{tab:compare}
\end{table*}

\subsection{Experimental Settings}
\textbf{Evaluation Setup.} Our experiments follow the experimental settings of prior methods~\cite{du2024demofusion, huang2024fouriscale, qiu2025freescale}, are built upon two publicly available foundational models: SDXL~\cite{podell2023sdxl} and SDXL-Turbo~\cite{sauer2024adversarial}, a distilled version optimized for few-step inference. We focus our evaluation on the demanding task of high-resolution synthesis, generating images at $2048\times2048$ and $4096\times4096$ to rigorously test its efficiency and quality. For quantitative evaluation, we utilize a set of 1000 prompts randomly sampled from the LAION/LAION2B aesthetic dataset to ensure a diverse and representative range of generated content.\\
\textbf{Baselines.} We compare proposed PixelRush with other tuning-free high-resolution image generation method (i) Direct inference at target resolution from SDXL~\cite{podell2023sdxl} (SDXL-DI) (ii) FouriScale~\cite{huang2024fouriscale} (iii) DemoFusion~\cite{du2024demofusion} (iv) FreeScale~\cite{qiu2025freescale}. Unlike other methods that apply the standard SDXL model across all stages, our proposed PixelRush leveraging the few-step SDXL-Turbo for upsampling.

\noindent\textbf{Evaluation Metrics.}~Our quantitative evaluation employs two standard metrics: Fréchet Inception Distance (FID)~\cite{heusel2017gans}, and Inception Score (IS)~\cite{salimans2016improved}. We compute FID and IS between our high-resolution generations and a dataset of real images.
\subsection{Qualitative Comparison}
To visually assess the performance of our method, we provide qualitative comparisons against state-of-the-art training-free approaches in Fig.~\ref{fig:compare}. We present results for both $2048 \times 2048$ and $4096 \times 4096$ resolution, with zoomed-in regions to highlight fine-grained details. The visual results corroborate our quantitative findings and reveal the distinct failure modes of prior methods, which PixelRush successfully overcomes.
\begin{itemize}
    \item \textbf{Naive Direct Inference.} SDXL-DI~\cite{podell2023sdxl} produces highly unnatural textures and significant object repetition. We attribute this problem to the diffusion model's inability to generalize to resolutions far beyond its training distribution~\cite{du2024demofusion, qiu2025freescale}.
    \item \textbf{Object Repetition.} DemoFusion~\cite{du2024demofusion} struggles to maintain global consistency, leading to structural artifacts like the duplicated dragon head in the top example. Our method avoids this by using a DDIM-inverted~\cite{song2020denoising} latent as a strong structural prior, ensuring the final image remains coherent.
    \item \textbf{Unnatural Textures.} Approaches that directly manipulate the frequency domain, such as FouriScale~\cite{huang2024fouriscale} and FreeScale~\cite{qiu2025freescale}, often introduce unwanted texture artifacts. FouriScale exhibits repetitive, grid-like patterns, while FreeScale suffers from excessive high-frequency noise, making details appear harsh and unnatural. In contrast, PixelRush synthesizes sharp yet natural-looking details that are consistent with the scene's context.
\end{itemize}
This visual evidence demonstrates that PixelRush not only avoids these common failure modes but also sets a new standard for sharpness and structural preservation in high-resolution synthesis.
\subsection{Quantitative Comparison}
We present a quantitative comparison of our method, PixelRush, against state-of-the-art training-free methods for 2K $(2048\times2048)$ and 4K $(4096\times4096)$ image generation in Table~\ref{tab:compare}. 
At the $2K$ resolution, PixelRush establishes a new state-of-the-art. It achieves an FID score of 50.13, significantly outperforming all baselines and notably surpassing the previous best method, FreeScale~\cite{qiu2025freescale} (52.87). Our method again leads with an IS of 14.32, a substantial improvement over FreeScale's 13.56. At $4K$ resolution, PixelRush maintains its superior quality, achieving the best FID score of 54.67 and the highest IS of 13.75.

The most remarkable aspect of these results is that PixelRush achieves this superior performance in a single sampling step. As shown in Tab.~\ref{tab:compare}, our PixelRush generates a $2K$ image in just 4 seconds, which is $10$ to $22$ faster than other tuning-free methods and $7$ times faster than direct inference. This gap widens at $4K$ resolution, where our method requires only 20 seconds. This represents a remarkable $12$ to $34$ acceleration over baselines that take several minutes (247-680s) to generate a single image. This fundamentally breaks the conventional trade-off between generation speed and perceptual quality. While existing methods require multiple steps, often sacrificing fidelity for speed, our approach demonstrates that it is possible to attain both state-of-the-art quality and extreme computational efficiency simultaneously.

\subsection{Ablation Study}
\subsubsection{Ablation on Base and Refinement Models}

We analyze the impact of model choice by testing various popular diffusion backbones for both the base and refinement stages in our pipeline (Table~\ref{tab:compare_ablation}). The results demonstrate that PixelRush is robust to the choice of model, achieving consistently high performance. This confirms the generalizability of our approach.

\begin{table}[t]
  \centering
  \resizebox{0.45\textwidth}{!}{
    \begin{tabular}{l l c c c}
      \toprule
      \multirow{2}{*}{Base} & \multirow{2}{*}{Refinement}
      & \multicolumn{3}{c}{$2048\times 2048$} \\
      \cmidrule(lr){3-5}
      & & FID ($\downarrow$) & IS ($\uparrow$) & Time (sec) \\
      \midrule
      SDXL~\cite{podell2023sdxl} & SDXL-turbo~\cite{sauer2024adversarial} & \textbf{50.13} & \textbf{14.32} & \textbf{4} \\
      SDXL~\cite{podell2023sdxl} & SD-turbo~\cite{sauer2024adversarial} & 52.75 & 13.83 & 4 \\ 
      SANA~\cite{xie2025sana} & SDXL-turbo~\cite{sauer2024adversarial} & 57.48 & 13.51 & 4 \\ 
      SDXL~\cite{podell2023sdxl} & Pixart-$\delta$~\cite{chen2024pixart} & 50.31 & 14.23 & 16 \\
      \bottomrule
    \end{tabular}
  }
  \caption{\textbf{Quantitative comparison between different configurations of base and refinement models}}
  \vspace{-5mm}
  \label{tab:compare_ablation}
\end{table}



\subsubsection{Ablation on PixelRush Components}
To quantify the contribution of each proposed component, we perform an incremental ablation study on 2K image generation. We begin with a baseline (A) that uses DDIM inversion to initiate and a standard 50-step full reverse diffusion process for refinement. From there, we progressively intergrate our contributions: the partial inversion scheme to improve efficiency, the usage of few-step model for further acceleration, our Gaussian filter blending to resolve artifacts, and finally the noise injection technique to tackle oversmoothing and arrive at the proposed PixelRush. The results are summarized in Table~\ref{tab:ablation_components}. Introducing partial inversion provides a $3.7\times$ speedup while also providing comparable quality, indicated by FID and IS score. Leveraging a few-step model achieves 4-sec inference time, but this extreme acceleration comes at the cost of reduced quality (FID increased to 57.23). This quality drop is then resolved by our subsequent components: the Gaussian filter blending addressed the checkerboard artifacts and noise injection counteracts the oversmoothing, signiticantly improving the FID.
\begin{table}[t]
  \centering
    \resizebox{.47\textwidth}{!}{
  \begin{tabular}{@{}l c c c c}
    \toprule
  Config &$\#$ steps & FID ($\downarrow$) & IS ($\uparrow$)  & Time (sec) \\
    \midrule
    Baseline (A) &50 &54.70 &13.92 &67 \\
    + Partial Inversion (Sec.~\ref{subsec:partial_gaussian}) &15 &52.90 &13.89 &18 \\
    
    + Few-step Model  (Sec.~\ref{subsec:speed-up})  &1 &57.23 &13.65 &4\\
    + Gaussian Blend (Sec.~\ref{subsec:feathering})  &1 &56.16 &13.77 &4 \\ 
    + Noise Injection (Sec.~\ref{subsec:noise_inject})  &1 &\textbf{50.13} &\textbf{14.32} &\textbf{4}\\ 
    \bottomrule
  \end{tabular}}
  \caption{\textbf{Ablation study on the components of PixelRush.}}
  \vspace{-3mm}
  \label{tab:ablation_components}
\end{table}
\subsubsection{Analysis of Partial Inversion Timestep}
\begin{table}[t]
  \centering
    \resizebox{.45\textwidth}{!}{
  \begin{tabular}{@{}l c c c c}
    \toprule
  Config &$\#$ steps & FID ($\downarrow$) & IS ($\uparrow$)  & Time (sec) \\
    \midrule
    Baseline (A) &50 &54.70 &13.92 &67 \\
    Ours ($K=249$)  &1 &\textbf{50.13} &\textbf{14.32} &\textbf{4} \\
    
    Ours ($K=499$)  &2 &66.24 &13..19 &7\\
    Ours ($K=749$)   &3 &72.34 &12.98 &10 \\ 
    Ours ($K=999$)   &4 &79.45 &12.01 &13\\ 
    \bottomrule
  \end{tabular}}
  \caption{\textbf{Analysis of Partial Inversion Timestep $K$ with a Few-Step Model}}
  \vspace{-4mm}
  \label{tab:ablation_timestep}
\end{table}

A core contribution of our work is adapting partial inversion to the few-step regime. We conducted an ablation on the intermediate step $K$. with results shown in Tab.~\ref{tab:ablation_timestep}. The table reveals that an optimal partial inversion timestep is a shallow inversion depth $K=249$. At this setting, our method is not only $16\times$ faster than 50-step baseline ($4$s versus 67s) and but also achieves significantly better FID score (50.13 versus 54.70). However, as we increase $K$ from 249, performance all degrades. We hypothesize this is due to an incompatibility between multi-step DDIM inversion and few-step models like SDXL-Turbo, which is shown in recently work~\cite{garibi2024renoise}. 
Therefore, the partial inversion enables us to find the principled and highly effective timestep $K$ that maximizes both speed and quality.
\section{Conclusion}
In this paper, we address the critical challenge of slow and impractical high-resolution image generation in pretrained diffusion models. We identified a fundamental inefficiency in existing training-free methods: their reliance on a full, multi-step denoising process is redundant for refinement. To resolve this, we introduced \textbf{PixelRush}, a novel pipeline built on four key contributions: a partial inversion technique that focuses computation exclusively on high-frequency detail synthesis, an integration with few-step diffusion models to markedly boost the refinement speed, a Gaussian smooth blending algorithm that eliminates patch boundary artifacts, and a noise injection technique to tackle the oversmoothing problem in the low-step regime. Our extensive experiments demonstrate that PixelRush achieves state-of-the-art quantitative and qualitative results while providing an unprecedented acceleration $10\times$ to $35\times$ over prior methods. This transforms high-resolution generation from a time-intensive offline task into a practical process.

{
    \small
    \bibliographystyle{ieeenat_fullname}
    \bibliography{main}
}
\clearpage
\appendix
\setcounter{page}{1}
\maketitlesupplementary

\paragraph{Overview.}
The supplementary material provides additional implementation details in Sec.~\ref{sec:supp_impl}, further quantitative results in Sec.~\ref{sec:supp_quan}, and an extended gallery of qualitative results in Sec.~\ref{sec:supp_qual}.
\section{Implementation Details}
\label{sec:supp_impl}
For the base generation stage, we use the base model SDXL to generate a base image at its native $1024\times1024$ resolution. We employ a classifier-free guidance scale of 7.5 for all experiments. The reverse process in our refinement stage uses deterministic DDIM sampling, with the DDIM eta parameter set to 0. The upsampling operator $\mathbb{UP}$ is implemented as standard bicubic interpolation performed in the pixel domain (RGB space). Our pipeline operates in a cascaded manner. For example, to generate an $8192\times8192$ image, the process starts with a $1024\times1024$ base image, which is then progressively upscaled and refined through $2048\times2048$ and $4096\times4096$ resolutions to reach the final target. For the patch extraction process, the latent is tiled into patches with a $50\%$ overlap along both spatial dimensions. For our noise injection technique, the spherical interpolation coefficient $\lambda$ from Eq.~\ref{eq:noise_injection} is set to a fixed value of 0.95 for all experiments.
\section{Additional Quantitative Results}
\label{sec:supp_quan}
\subsection{Comparison with Super-Resolution}
As demonstrated in prior work~\cite{du2024demofusion, qiu2025freescale}, traditional super-resolution methods have been shown to underperform compared to training-free diffusion pipelines like FreeScale~\cite{qiu2025freescale} and DemoFusion~\cite{du2024demofusion}. Consequently, we exclude these methods from our baselines to focus on the state-of-the-art training-free paradigms.
\subsection{Additional Metrics}
\begin{table}[t]
  \centering
    \resizebox{.48\textwidth}{!}{
  \begin{tabular}{@{}l c c c c}
  \toprule
        
      & \multicolumn{2}{ c }{$2048\times2048$} &\multicolumn{2}{ c }{$4096\times 4096$}\\
    \cmidrule(lr){2-3}\cmidrule(lr){4-5}
& $\text{FID}_c$ ($\downarrow$) & CLIP ($\uparrow$)   & $\text{FID}_c$ ($\downarrow$) & CLIP ($\uparrow$)   \\
    \midrule
    SDXL-DI~\cite{podell2023sdxl}&47.84 &31.37  &73.96 &26.87 \\
    FouriScale~\cite{huang2024fouriscale} &51.90 &28.27  &69.56 &29.97  \\ 
    DemoFusion~\cite{du2024demofusion} &43.39 &29.45  &52.14 &29.01 \\ 
    FreeScale~\cite{qiu2025freescale} &29.64 &33.21  &34.01 &33.10 \\ 
    PixelRush\textsuperscript{*} & \textbf{29.13} &$\textbf{33.29}$  & $\textbf{32.83}$ &$\textbf{33.23}$ \\
    \bottomrule
  \end{tabular}  }
  \caption{\textbf{Additional quantitative comparison using $\text{FID}_c$ and CLIP score.} The results confirm that PixelRush consistently outperforms all baselines on both realism ($\text{FID}_c$) and text-prompt alignment (CLIP) at $2K$ and $4K$ resolutions.}
  \vspace{-1mm}
  \label{tab:supp_compare}
\end{table}

To further validate our findings, we conduct an extended quantitative analysis. A key drawback of standard FID is the need to resize images to resolution $299 \times 299$ penalizes high-resolution methods by discarding the very details they are designed to synthesize. Inspired by~\cite{qiu2025freescale}, we report $\text{FID}_c$, a version of FID computed using local crops without resizing for better evaluating the image detail construction. Note that $\text{FID}_c$ scores might be lower than FID scores due to the increase in the number of samples. Furthermore, to evaluate text-prompt alignment, we include the CLIP score (ViT-B/32). As demonstrated in Tab.~\ref{tab:supp_compare}, PixelRush consistently outperforms all competing methods.
\subsection{Reducing Overlap Fraction To Speed Up}
\begin{table}[t]
  \centering
    \resizebox{.47\textwidth}{!}{
  \begin{tabular}{@{}l c c c c}
    \toprule
  \multirow{2}{*}{Overlap Fraction} & \multirow{2}{*}{$\#$ patches}
      & \multicolumn{3}{c}{$4096\times 4096$} \\
      \cmidrule(lr){3-5}
      & & FID ($\downarrow$) & IS ($\uparrow$) & Time (sec) \\
    \midrule
    0.5 &49 &54.67 &13.75 &20 \\
    0.25  &25 &54.28 &13.71 &16 \\
    \bottomrule
  \end{tabular}}
  \caption{\textbf{Analysis of the patch overlap fraction on 4K image generation.} A smaller overlap of $25\%$ significantly reduces the number of patches, leading to faster inference}
  \label{tab:supp_overlap_size}
\end{table}
\begin{figure}[t]
\centering
\includegraphics[width = 0.48\textwidth]{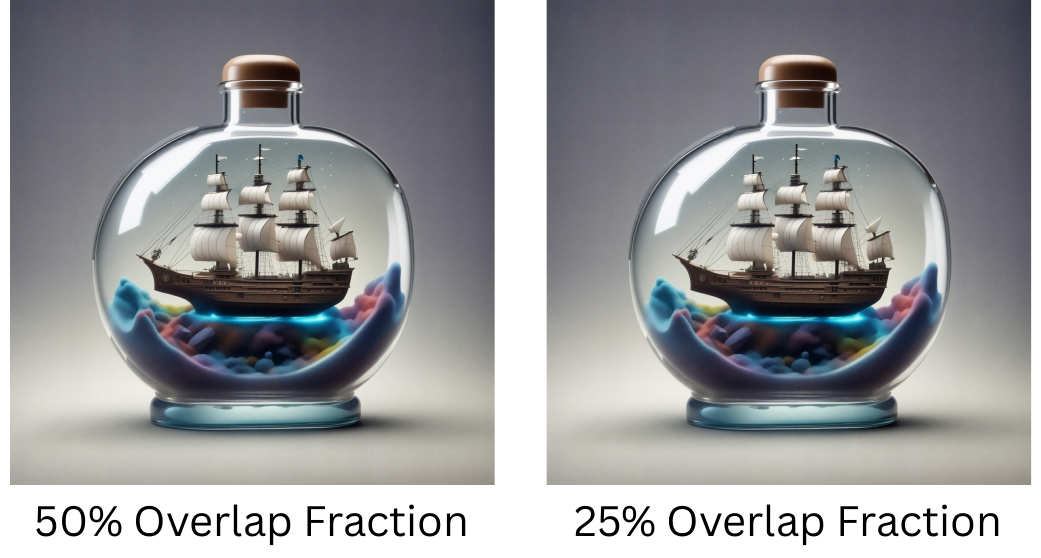}
\caption{\textbf{Qualitative comparison of different patch overlap fractions.} This comparison shows that reducing the patch overlap from $50\%$ to $25\%$ yields a speedup and produces a visually indistinguishable result.}
\label{fig:supp_reduce_patch}
\end{figure}
Following prior work~\cite{du2024demofusion}, we partition the latent space into patches with an overlap of $50\%$. We note that the necessity for such a large overlap in previous methods often stems from their use of stochastic $q$-sampling, which requires a wide blending region to average out inconsistencies. In contrast, our use of deterministic DDIM-inv preserves the global structure better. This is a potential for optimization, as a smaller overlap could potentially increase inference speed without a significant loss in quality.

The number of patches can be quantified. The number of patches $n$ required along one spatial dimension is related to the upscale factor $M$, and the overlap fraction $o$ by the formula:
\begin{equation}
    n - (n-1) \cdot o = M
\end{equation}
For example, to generate a $4096 \times 4096$ image ($M=4$), a $50\%$ overlap ($o = 0.5$) requires $n=7$, for a total of $n^2=49$ patches. However, the formula shows that when using a $25\%$ overlap, we just need $n=5$, reducing the number of patches to only $25$ patches. As shown empirically in Tab.~\ref{tab:supp_overlap_size} and Fig.~\ref{fig:supp_reduce_patch}, reducing the patch count is an  optimization for our pipeline, yielding a faster inference process with no loss in perceptual quality.

\subsection{Ablation On Noise Injection}
\begin{table}[t]
  \centering
    \resizebox{.43\textwidth}{!}{
  \begin{tabular}{@{}l c c c c}
    \toprule
  \multirow{2}{*}{$\lambda$} & \multirow{2}{*}{$\#$ steps}
      & \multicolumn{3}{c}{$2048\times 2048$} \\
      \cmidrule(lr){3-5}
      & & FID ($\downarrow$) & IS ($\uparrow$) & Time (sec) \\
    \midrule
    0.7 &1 &58.57 &13.72 &4 \\
    0.8  &1 &52.57 &14.09 &4 \\
    
    0.9  &1 &50.02 &14.15 &4\\ 
    0.95 &1 &50.13 &14.32 &4 \\
    \bottomrule
  \end{tabular}}
  \caption{\textbf{Analysis of the noise injection coefficient $\lambda$.} We analyze the sensitivity of our method to the choice of $\lambda$ for $2K$ image generation. We identify $\lambda=0.95$ as the optimal setting, yielding the best FID and IS scores.}
  \label{tab:supp_noise_injection}
\end{table}

To analyze the sensitivity of our method to the noise injection coefficient $\lambda$, we performed an ablation across a range of values. The results presented in Tab.~\ref{tab:supp_noise_injection} identify $\lambda=0.95$ as the optimal setting, which we use throughout our work. 
\section{Additional Qualitative Results}
\label{sec:supp_qual}
\subsection{Extension To High-Resolution Video Generation}
\begin{figure}[t]
\centering
\includegraphics[width = 0.49\textwidth]{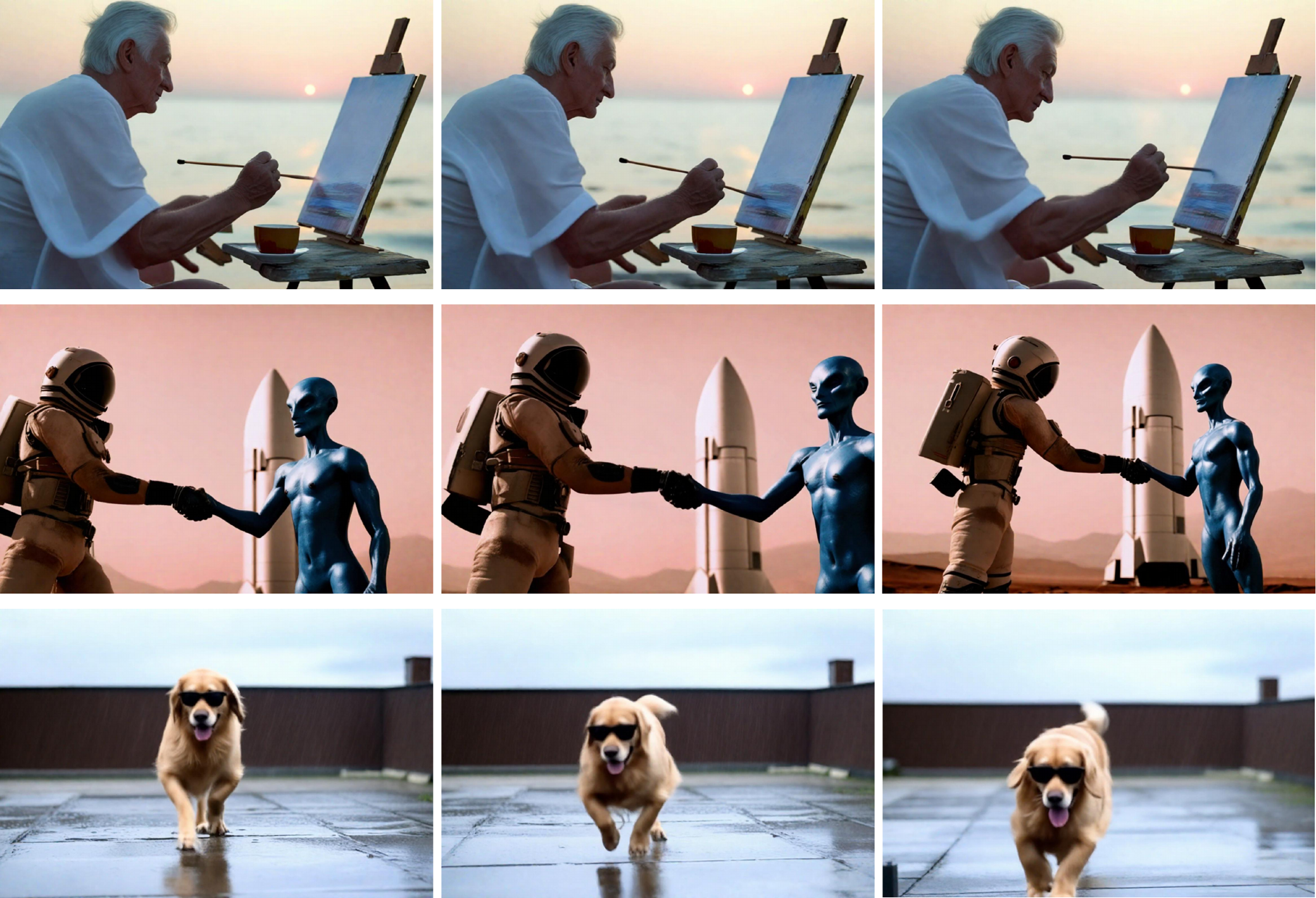}
\caption{\textbf{Extending PixelRush to high-resolution video synthesis.} The figure showcases consecutive frames from videos upscaled frame-by-frame with our method. PixelRush successfully enhances the detail and sharpness in each frame, demonstrating its applicability to video content. Best viewed \textbf{ZOOMED-IN}.}
\label{fig:supp_vid}
\end{figure}
Our method can be directly extended to high-resolution video generation by applying it frame-by-frame. We apply PixelRush independently to each frame of low-resolution video generated by CogVideoX-5B. As demonstrated in Fig.~\ref{fig:supp_vid}, this approach successfully enhances the resolution and synthesize fine-grained details within each individual frame. However, because our method is applied on a per-frame fashion, it does not enforce temporal consistency across the video sequence, could result in flickering artifacts between frames. Improving temporal coherence presents a potential direction for future research.




\end{document}